\documentclass[twocolumn]{article}

\usepackage[width=17cm,height=22cm]{geometry}
\usepackage[french, english]{babel}
\usepackage[utf8]{inputenc}
\usepackage{fancyvrb}
\usepackage{authblk}
\usepackage{hyperref}
\usepackage{amsfonts}
\usepackage{amssymb}
\usepackage{amsthm,amsmath}
\usepackage{hyperref}

\usepackage{graphicx}
\usepackage{comment}
\usepackage{stfloats}

\title{Autoencoder-based time series clustering with energy applications}
\author[1]{Guillaume Richard}
\author[1]{Benoît Grossin}
\author[1]{Guillaume Germaine}
\author[1]{Georges Hébrail}
\author[1]{Anne de Moliner}
\affil[1]{EDF R\&D}

\date{}

\begin{document}
\maketitle

\begin{abstract}
Time series clustering is a challenging task due to the specific nature of the data. Classical approaches do not perform well and need to be adapted either through a new distance measure or a data transformation. In this paper we investigate the combination of a convolutional autoencoder and a k-medoids algorithm to perfom time series clustering. The convolutional autoencoder allows to extract meaningful features and reduce the dimension of the data, leading to an improvement of the subsequent clustering. Using simulation and energy related data to validate the approach, experimental results show that the clustering is robust to outliers thus leading to finer clusters than with standard methods.
\end{abstract}

\medskip


\section{Introduction and related work}
\label{sec:lentete}

Time series clustering is a significant problem in time series data mining. The goal is to group similar time series into the same clusters. It allows to identify different structures in the dataset as an exploration tool that can be then used for other tasks such as summarization, classification, visualization ... Applications in various fields (medical, financial services, engineering, ...) have been developed in recent years. Only whole time series clustering will be considered here as opposed to time point or subsequence clustering.\\
Clustering techniques have been developed for a long time and different algorithms have been widely used: k-means, hierarchical clustering, self-organizing maps, ... Time series are high-dimensional and subject to noise. Data points are highly correlated which can lead those algorithms to perform poorly. Different extensions have been proposed to adapt clustering algorithms to time series data \cite{Aghabozorgi}. They can be separated into two main categories: representation-based or similarity-based. \\
Similarity-based approaches try to find a good way to measure the distance between time-series. As the euclidean distance is generally not well suited for time series analysis different similarities have been designed: Dynamic Time Warping (DTW) \cite{Sakoe:1990:DPA:108235.108244}, Longest Common Sub-Sequence (LCSS), MINDIST, Edit Distance with Real Penalty (ERP), ... These distances try to challenge issues such as temporal shift, noise or offsets. \\
Representation-based methods aim primarily to reduce the dimension of time series. The main goal is to extract the underlying structure of the time series and remove noise or other effects. The choice of the feature extraction method is then the main factor for the final clustering. Time series analysis has led to the development of various representations: Discrete Fourier Transform (DFT), Discrete Wavelet Transform (DWT), Principal Component Analysis (PCA), Piecewise Aggregate Approximation (PAA), Symbolic Approximation (SAX), ... which have been used as a pre-processing tool for clustering \cite{Vlachos03awavelet-based} \cite{cao2015time}. \\
This work falls within the representation-based methods. Artificial neural networks have been very successful during the last decade in many fields such as computer vision or natural language processing. Autoencoders are a class of neural networks intended to learn compressed representations of the data in an unsupervised setting. They produce a non-linear transformation to a smaller latent space through an encoder. A decoder is then used to reconstruct the data based on the latent representation. Convolutional autoencoders use convolutional operators to perform the feature extraction. It is well suited to temporal data since it allows local shift-invariance and captures the shape of the time series.\\
Most previous unsupervised neural network for time series modeling used recurrent architectures and few used convolutional ones. Using a convolutional autoencoder has been proposed for pattern discovery \cite{bascol:hal-01374576}. In a classification framework various convolutional architectures have been designed \cite{Cui2016MultiScaleCN}. In the computer vision community convolutional clustering has been proposed, including clustering in the learning phase of the convolutional autoencoder \cite{Guo2017DeepCW} \cite{DBLP:journals/corr/DundarJC15}. To the best of the authors' knowledge no convolutional autoencoder has been used for time series clustering.\\
This paper is divided into the following sections: firstly the method is presented and gives a metric for outliers in a time series dataset. Then experiments are conducted on real and synthetic data: a comparative study with other feature extraction approaches shows that our method gives good results in term of robustness to outliers.

\section{Proposed method}
\label{sec:Method}
\begin{figure}[htbp]
  \centering
  \includegraphics[scale=1]{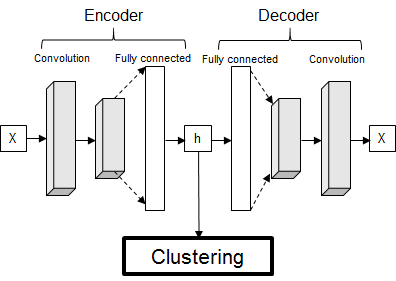}
  \caption{Proposed two-stage model: convolutional autoencoder and clustering}
  \label{fig:proposed_model}
\end{figure}

In this section we describe the time series clustering model in details. As shown in Figure \ref{fig:proposed_model} the clustering is done in two steps. First a convolutional autoencoder is trained to map the input time series to a latent vector which is then used to reconstruct the input. The latent vector is then used as an input for a clustering algorithm. We also discuss about the definition of outliers in a time series dataset.

\subsection{Convolutional autoencoders}

A convolutional autoencoder (CAE) is a special type of autoencoder that is made of one or more convolutional layers. It maps an input time series $x \in \mathbb{R}^T$ to a latent vector $h \in \mathbb{R}^{m}$ with $m<T$. This is part of the \textit{Load2Vec} approach which aims at finding meaningful representations of eletric power consumption time series (load curves).\\
The encoding function is a composition of several layers: convolutional and fully connected layers. A fully connected layer takes an input $z \in \mathbb{R}^n$ and transforms it into an output $y \in \mathbb{R}^l$ through an activation function $\sigma$ :
\begin{equation}
\label{eq:fc}
y = \sigma(Wx+b) 
\end{equation}

\noindent where $W \in \mathbb{R}^{l \times n}$ and $b \in \mathbb{R}^l$ are respectively the weight matrix and the bias of the layer to be learned. \\
Similarly a convolutional layer is defined by a transformation :  
\begin{equation}
\label{eq:conv}
y_i = \sigma(\sum_{j=1}^r W_j x_{1+(i-1)s+j} + b_j) 
\end{equation}
\noindent where $W \in \mathbb{R}^{r}$ and $b \in \mathbb{R}^r$ are respectively the kernel matrix and the bias of the layer to be learned with kernel size $r$ and stride $s$. The first part of the encoder is composed of convolutional layers before a fully connected part for the latent vector. \\
The decoder is composed of a fully connected layer followed by deconvolution layers. A deconvolution layer is the invert operator of a convolution layer. It can be seen as a convolution layer but with an upsampling factor.\\
Finally the output of the decoder is a vector $\hat{x} \in \mathbb{R^T}$. We then define a loss function $\mathcal{L}$ which measures the similarity between $x$ and $\hat{x}$. The objective of the autoencoder is to learn the optimal parameters of the layers of the network to minimize values of objective function:
\begin{equation}
\min \sum_{i=1}^N \mathcal{L}(x^i,\hat{x}^i)
\end{equation}

The main constraint of the convolutional autoencoder is the size of the latent vector $m$. We also add a penalization term with the $l^2$ norm on the layer weights of the layers to avoid high weight parameters. The parameters are learned using a stochastic gradient descent algorithm or one of its variants.

\subsection{Clustering}

The clustering step is performed through a K-Medoids algorithm on the latent vector. Since the K-Medoids algorithm is sensible to the scale of each dimension, the latent vector has to be normalized. The two main advantages of K-Medoids are its simplicity and its robustness to outliers.\\
It is worth noting that another possibility would have been to regularize the latent vector in the learning of the autoencoder using a variational autoencoder \cite{journals/corr/KingmaW13}. In our trials the autoencoder ended up by being over-regularized, even using annealing strategies so we did not use that regularization.\\
To choose the number of clusters we use the elbow method. The feature extraction is generic and Other clustering techniques could be used such as agglomerative clustering or K-means. 

\subsection{Outliers definition}

One of the main issues of traditional clustering techniques is the robustness to outliers. But how to define an outlier in a dataset is a challenging task. Different approaches have been proposed: Mahalanobis distance, Outlier Detection using Indegree Number, ...  In this paper we use the Local Outlier Factor metrics\cite{Breunig:2000:LID:335191.335388}.\\
To define this metric we have to define a distance on our initial space. Here we choose the euclidean distance.\\
A \textit{k-NN} graph is computed on the dataset. Defining the \textit{local reachibility density} of a sample $x_i$ as : $$lrd(x_i)=\left( \frac{\sum\limits_{x_j \in \mathcal{N}(x_i)} \max(d_k(x_j), d(x_i, x_j))}{k} \right) ^{-1}$$
where $d_k(x_j)=\max_{x_k \in \mathcal{N}(x_i)} d(x_j, x_k)$ and  $\mathcal{N}(x_i)$ is the neighbourhood of $x_i$\\
The \textbf{Local Outlier Factor} is defined as: 

\begin{equation}
LOF(x_i)=\frac{\sum\limits_{x_j \in \mathcal{N}(x_i)} lrd(x_j)}{k lrd(x_i)}
\end{equation}

The main idea behind this metric is that it allows to define outliers locally. Therefore points that would be defined as outliers with other methods such as Mahalanobis because they belong to a small or sparse cluster will be considered as normal with this metric.\\ 
A threshold is manually defined such that every point with a LOF higher than the threshold is considered as an outlier. We use this definition to determine the number of outliers per cluster which quantifies the \textit{outlyingness} of a cluster.

\section{Experiments}
\label{sec:experiments}

\subsection{Settings}

We evaluated the proposed approach on both synthetic and real data. The synthetic dataset is generated using expertise on electricity consumption behaviours. The real dataset comes from the CER Smart Metering Project which monitored the consumption of 3,174 Irish homes and businesses during 2009 and 2010.\\
We compare several feature extraction approaches as a pre-processing for the K-medoids algorithm or different time series clustering algorithm :
\begin{itemize}
	\setlength\itemsep{0em}
	\item \textbf{K-medoids: } K-medoids performed on raw data (divided by the mean)
    \item \textbf{PCA+K-medoids:} only the 20 first components of PCA are used
    \item \textbf{Haar+IK-means:} the coefficients of a Discrete Wavelet Transform with Haar wavelet are used in an interactive K-Means as defined in \cite{Vlachos03awavelet-based}
    \item \textbf{DTW+K-medoids:} K-medoids based on Dynamic Time Warping similarity
    \item \textbf{CAE+K-medoids:} K-medoids performed on latent vector from the convolutional autoencoder
\end{itemize}

The used architecture for the CAE is the following for both experiments:

\begin{itemize}
	\setlength\itemsep{0cm}
	\item Conv1D: $64$ filters of length $3$, stride $2$, activation \textit{elu}
    \item BatchNormalization
    \item Conv1D: $128$ filters of length $5$, stride $2$, activation \textit{elu}
    \item BatchNormalization
    \item Flatten
    \item FullyConnected: $100$ units, activation \textit{elu}
    \item FullyConnected: $20$ units, activation \textit{linear}
    \item FullyConnected: $128 \times 96$ units, activation \textit{elu}
    \item BatchNormalization
    \item DeConv1D: $128$ filters of length $5$, activation \textit{elu}
    \item BatchNormalization
    \item DeConv1D: $64$ filters of length $3$, activation \textit{elu}
    \item Conv1D: $1$ filters of length $3$, activation \textit{tanh}
\end{itemize}

We include BatchNormalization layers for faster learning. Each implementation has been made using Python with Keras and Tensorflow for the convolutional autoencoder. 

\subsection{Synthetic data}

Due to the imperfect definition of outliers in a real dataset we validate the method using synthetic data. Using knowledge on traditional electricity consumption behaviour of residents and SMEs we simulate a synthetic dataset consisting of 2,000 time series.\\
The time series have a length of 384 with each point representing the daily consumption. Three classes are created: residential, SMEs and outliers with respectively 1470, 490 and 40 elements. 

\begin{itemize}
	\item \textbf{Residential clients:} those clients are characterized by a global trend depending on the date of the year (summer vs winter) and a slightly higher consumption during the weekend. Vacation periods are also added. 
    \item \textbf{SMEs:} the global trend is smaller than the one of residential clients and the consumption during the weekend is much smaller. 
    \item \textbf{Outliers:} those clients are either characterized as secondary homes (with non null consumption only on some weekends or vacation) or have random values. 
\end{itemize}

\begin{figure}[htbp]
  \centering
  \includegraphics[scale=0.33]{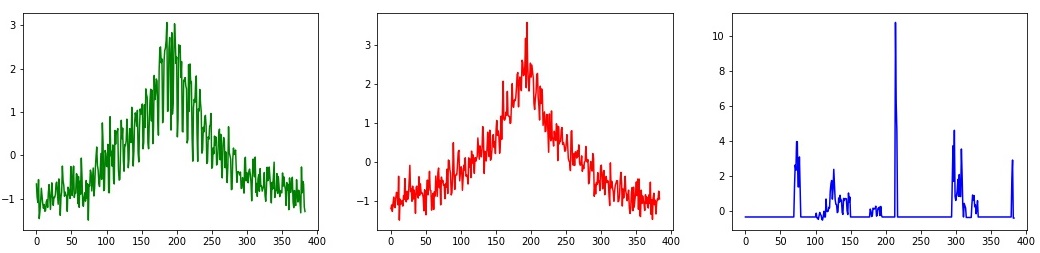}
  \caption{Green: SME ; Red: Residential ; Blue: Outlier}
  \label{fig:clusters_outliers}
\end{figure}

\begin{figure}[htbp]
  \centering
  \includegraphics[scale=1]{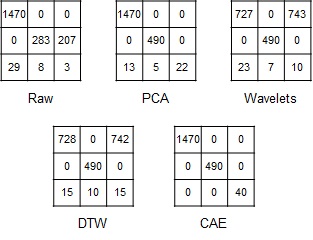}
  \caption{Confusion matrix between true labels and obtained clusters}
  \label{fig:results_synthetic}
\end{figure}

On Figure \ref{fig:results_synthetic} we compare the results of each approach. We present the number of elements of the three classes in each cluster: first line is the number of residential in each cluster, second line is the number of SME and the last line stands for outliers.

Our method separates perfectly the different classes in an unsupervised setting. PCA is also able to separate residentials from SMEs but suffers from some outliers. Other methods do not isolate an outlier class and clusters the two classes into three clusters by splitting one class in two parts.\\

\subsection{CER Smart Metering Project}

The approach was also tested using the electricity consumption from 3,174 \footnote{we only kept clients for which survey data was available} Irish households and SMEs between 2009 and 2010 \cite{CER}. Original data is sampled at a 30 minute step but were aggregated by day for dimension reduction and we consider only the first 384 days of the experiment. \\
As we want to differentiate clients by their behaviour and not their consumption level the time series are normalized client by client by dividing by the average consumption of the client. \\
Survey data is also available, mainly indicating if the client is a household or a small business. One can see that SMEs are characterized by a lower consumption during weekend whereas households have a slightly higher consumption on those days. The other main trend is the difference between summer and winter.\\
The pre-processing and subsequent clustering is very quick to compute: few seconds for raw data, PCA and Wavelets. The convolutional autoencoder takes 2 minutes to train working with a Quadro P6000. Finally the full computation of \textit{dtw} similarity on the whole dataset takes 10 minutes.

\begin{figure}[htpb]
  \centering
  \includegraphics[scale=0.4]{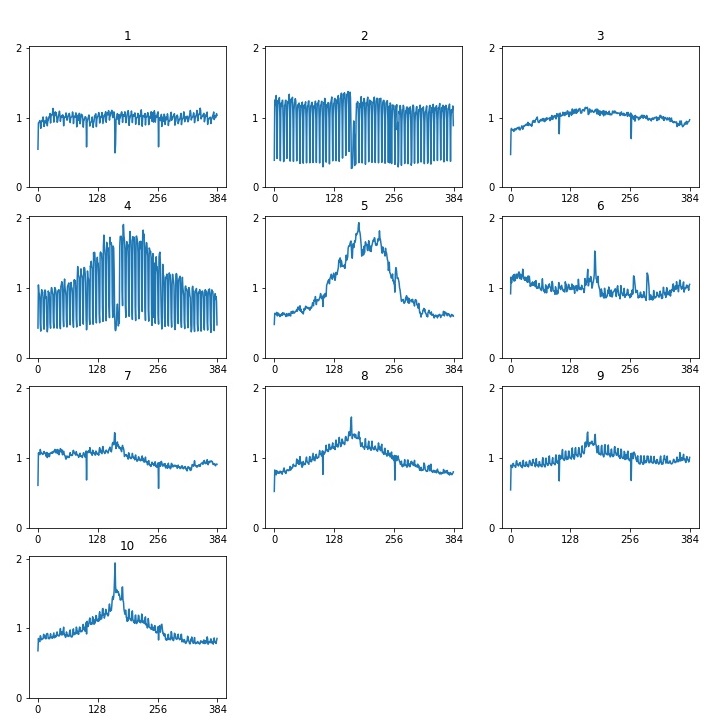}
  \caption{Centroids of each cluster found with the CAE+K-Medoids method}
  \label{fig:cluster_centers}
\end{figure}

\begin{figure*}[t]
  \centering
  \includegraphics[scale=0.7]{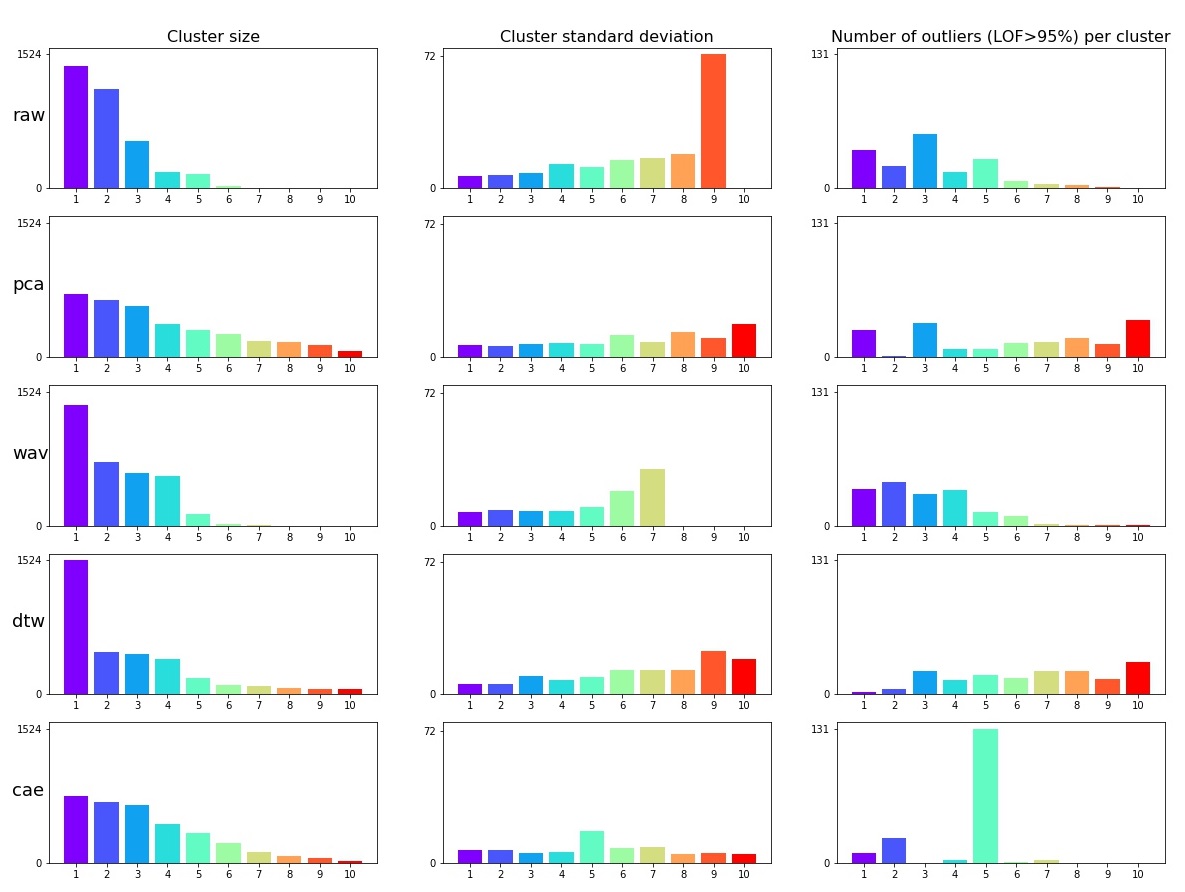}
  \caption{Left: number of elements per clusters ; Center: number of outliers in each cluster (an element is an outlier if its Local Outlier Factor is above the 95 \% quantile) ; Left: standard deviation of each cluster}
  \label{fig:clusters_outliers}
\end{figure*}

Using the elbow method we find that a good number of clusters would be 10 or 16. On Figure \ref{fig:clusters_outliers} are plotted the size of each cluster and the number of outliers per cluster. An outlier is defined by its Local Outlier Factor and here we set the number of outliers at $5\%$ of the dataset.\\
One can see that most methods do not cope well with outliers: either the outliers are spread between clusters or they are all gathered in the major class. But the proposed method allows to isolate a class with high dispersion which gathers the outliers. Then the rest of the data can be clustered into finer clusters.\\
Clusters have an interpretation and the centroids are shown in \ref{fig:cluster_centers}. Two clusters of SMEs come out and residential clients are clustered depending on other features: the peak in consumption during the 2010 winter or Christmas for example.\\
The outlier cluster is characterized by many clients with abnormal consumption patterns (often very small or with few very high values).\\
Tests were done with other number of clusters and the isolation of one \textit{outlier class} remains. After 20 clusters the class is split into two. Experimentations have also been conducted with K-Means and Self-Organizing Maps (SOM). Similar results were obtained with the CAE whereas performance with other feature extraction  dropped. This tends to show that convolutional autoencoder is a feature extractor for time series with very nice properties.

\section{Conclusion}

We propose a convolutional autoencoder as a feature extraction method for time series data. Experiments show that it allows to learn meaningful features to perform subsequent clustering. This approach performs well in terms of quality of clustering and robustness to outliers. It avoids the drawback of outlier removal from the dataset by gathering them into a single class.\\
Our method is efficient for several clustering algorithms. Furthermore extensions to the multivariate case can be done easily using the different dimensions as channels in the autoencoder. Including the clustering in the learning as in \cite{Guo2017DeepCW} did not yield the same improvement in our experiments.

\bibliography{biblio.bib}

\end{document}